\documentclass[10pt,twocolumn,letterpaper]{article}

\usepackage{3dv}
\usepackage{times}
\usepackage{epsfig}
\usepackage{graphicx}
\usepackage{gensymb}
\usepackage{amsmath}
\usepackage{amssymb}
\usepackage{authblk}
\usepackage{url}
\usepackage{balance}
\usepackage{lipsum}

\newcommand\blfootnote[1]{%
  \begingroup
  \renewcommand\thefootnote{}\footnote{#1}%
  \addtocounter{footnote}{-1}%
  \endgroup
}

\usepackage[pagebackref=true,breaklinks=true,letterpaper=true,colorlinks,bookmarks=false]{hyperref}

\threedvfinalcopy 

\def\threedvPaperID{54} 

\ifthreedvfinal\fi
\begin{document}

\title{A Unified Point-Based Framework for 3D Segmentation}

\author[1]{Hung-Yueh Chiang}
\author[2]{Yen-Liang Lin}
\author[1]{Yueh-Cheng Liu}
\author[1]{Winston H. Hsu}

\affil[1]{National Taiwan University}
\affil[2]{A9/Amazon}

\maketitle

\begin{abstract} \label{abstract}
3D point cloud segmentation remains challenging for structureless and textureless regions.
We present a new unified point-based framework\blfootnote{Code is available at \url{https://github.com/ken012git/joint_point_based}} for 3D point cloud segmentation that effectively optimizes pixel-level features, geometrical structures and global context priors of an entire scene.
By back-projecting 2D image features into 3D coordinates, our network learns 2D textural appearance and 3D structural features in a unified framework.
In addition, we investigate a global context prior to obtain a better prediction.
We evaluate our framework on ScanNet online benchmark and show that our method outperforms several state-of-the-art approaches.
We explore synthesizing camera poses in 3D reconstructed scenes for achieving higher performance.
In-depth analysis on feature combinations and synthetic camera pose verify that features from different modalities benefit each other and dense camera pose sampling further improves the segmentation results.
\end{abstract}

\section{Introduction} \label{sec:introduction}
Deriving a 3D map with high-level semantics is the key for intelligent navigation systems to interact with humans in the environment. 
A significant research effort has been invested in 3D classification and segmentation tasks. 
ScanNet \cite{dai2017scannet} and Matterport \cite{chang2017matterport3d} collect large-scale RGB-D video datasets, and provide semantic and instance annotations for 3D point clouds. 
Many following works \cite{dai20183dmv, wang2018sgpn, tchapmi2017segcloud, qi2017pointnet, qi2017pointnet++, su2018splatnet, liang2018deep,graham20183d,graham2015sparse} investigate various deep learning algorithms for 3D scenarios. 
Su et al.~\cite{su2015multi} obtain 3D representations by applying CNNs on 2D rendering images and aggregating multi-view features for 3D classification tasks.
Leveraging the success of CNN on the image pixel grid, 3D voxel networks, such as \cite{zhou2017voxelnet,maturana2015voxnet,graham20183d,graham2015sparse}, learn spatial relationships from discretized space by 3D convolution kernels. 
However, voxelization may bring quantization artifacts, which limits its utility to low-resolution point cloud data. 
Point-based networks \cite{qi2017pointnet, qi2017pointnet++} are proposed to alleviate the problem and directly process on input points, which are more efficient to represent geometry and flexible to different data formats such as depth sensor and lidar data.
However, most of them only use geometric features without considering the features from other modalities, e.g., image features.
In addition, the global context has been shown effective to 2D scene parsing tasks \cite{ParseNet, zhao2017pyramid}, but it is not yet investigated in recent 3D architectures.

\begin{figure}[t!]
\centering
\includegraphics[width=0.5\textwidth]{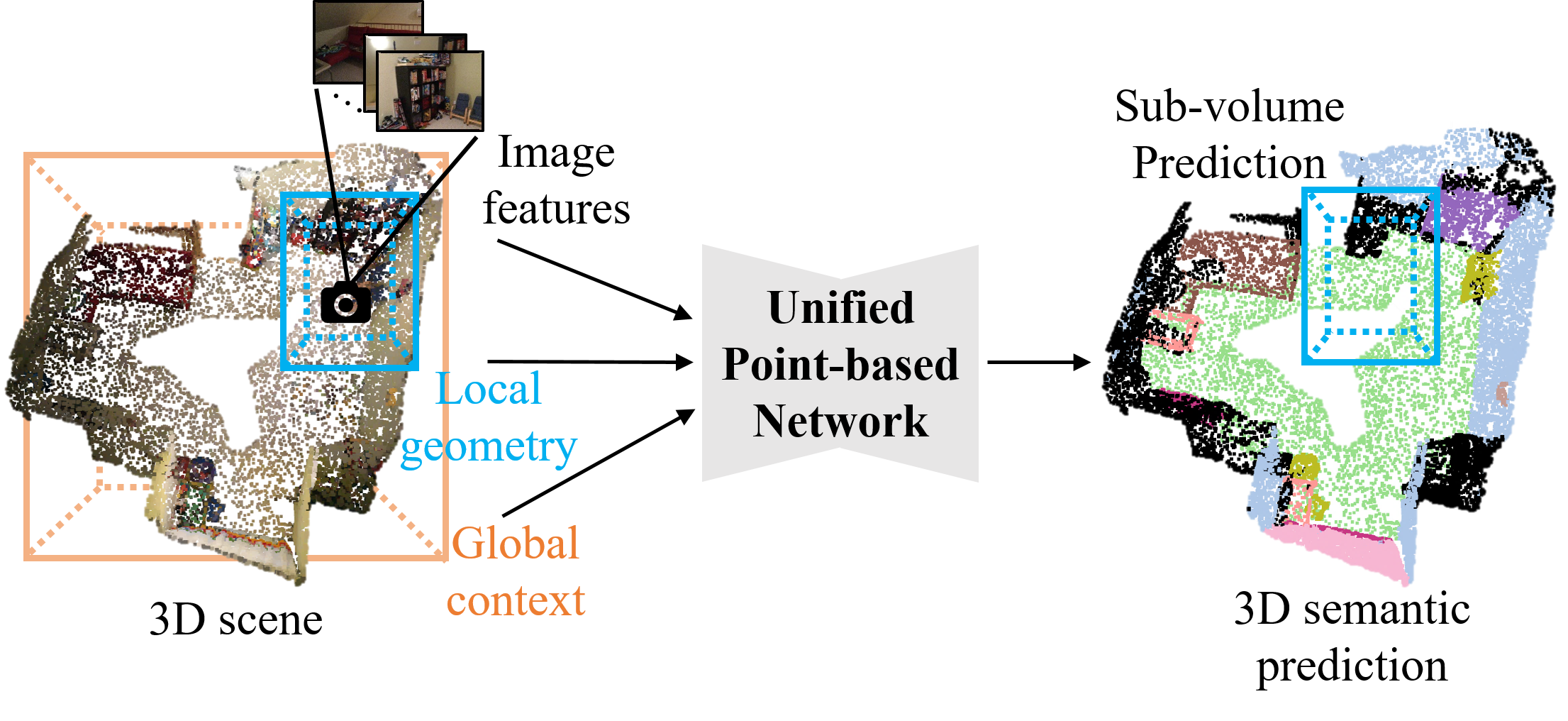}
\caption{
We propose to effectively optimize image features, geometrical structures and global context priors in a unified point-based framework.
} 
\label{fig:idea}
\end{figure}

\begin{figure*}[t!]
\centering
\includegraphics[width=\textwidth]{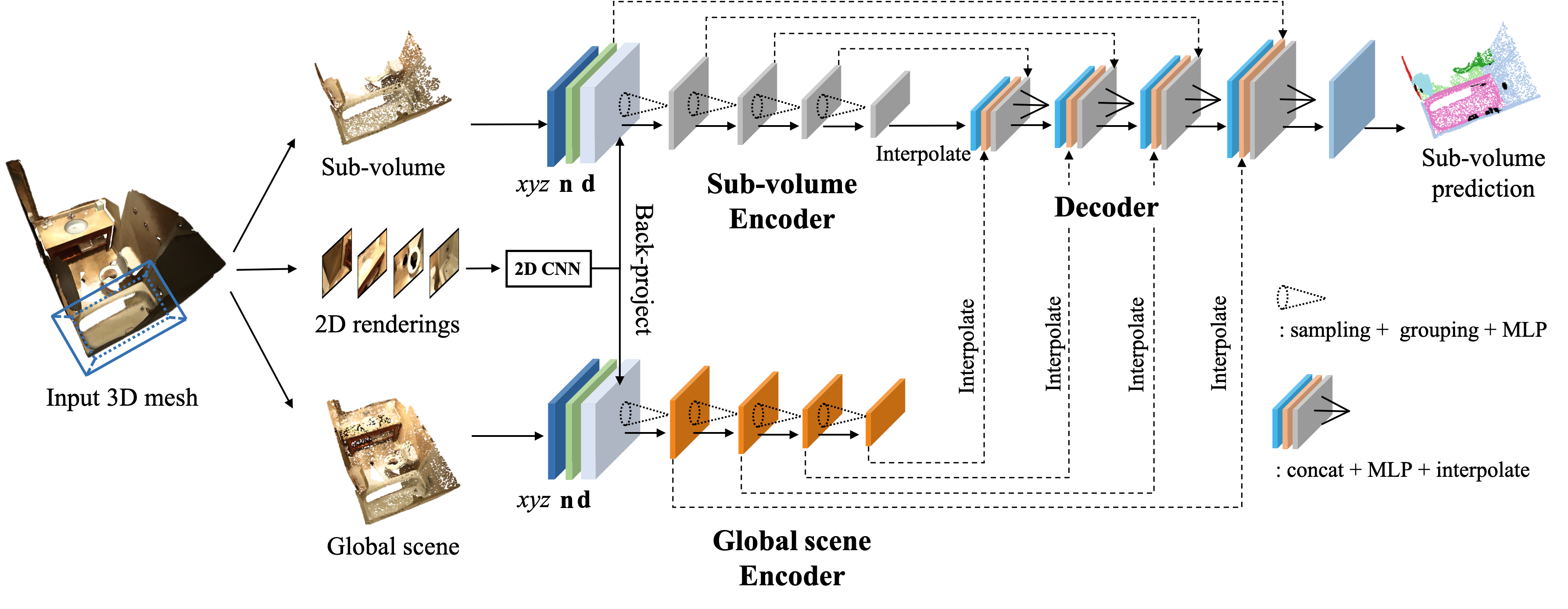}
\caption{
System overview of the proposed framework. 
We extract image features by applying a 2D segmentation network and back-project (cf. Figure \ref{fig:backproject}) the 2D image features into 3D space.
Both 2D image features and 3D point features are concatenated in the same point coordinates. 
The sub-volume and scene encoder extract the features for local details and global context priors respectively. 
The decoder fuses the local and global features and generate the semantic predictions for each point in the sub-volume. 
}   
\label{fig:network structure}
\end{figure*}

We propose a unified point-based framework for 3D point cloud segmentation, as shown in Figure \ref{fig:network structure}, that effectively leverages 2D pixel-level image features, 3D point-level structures, and global contexts priors within a scene.
%
%
The experimental results show that 2D, 3D, and global context features benefit each other (Table \ref{tb:features}).
%
%
To improve the wrongly estimated camera pose from structure from motion, we explore synthetic camera poses in 3D scenes.
The result shows that synthetic camera pose sampling further improves our performance on ScanNet testing set from $62.1\%$ to $63.4\%$ (Table \ref{tb:test set}).
Our unified framework demonstrates superior performance over several state-of-the-art methods (Table \ref{tb:test set}): $63.4\%$ (ours) vs. $48.4\%$ (3DMV) \cite{dai20183dmv}, and $39.3\%$ (SplatNet) \cite{su2018splatnet}.

The main contributions of this work conclude as following:
\begin{itemize}
    \item We propose to effectively leverage 2D image features, geometric structures and global context priors within an entire scene into a unified point-based framework, which is shown to experimentally outperform several state-of-the-art approaches on ScanNet benchmark \cite{ScanNetBenchmark}.
    \item We provide an in-depth analysis of various decision choices (e.g., point features, sub-volume strides, synthetic camera models) of our framework to achieve better performance.
    \item Through experimenting on different feature combinations, we demonstrate the semantic segmentation is improved by textural, geometric and global context information.
\end{itemize}

\section{Related Work} \label{sec:related work}
Some works \cite{wang2018sgpn, liu2018floornet, liang2018deep, su2018splatnet, dai20183dmv} leverage both structural and textural information on different 3D tasks. 
3D semantic segmentation can be categorized into image-based \cite{hermans2014dense, mccormac2017semanticfusion}, voxel-based \cite{dai2017scannet, dai20183dmv, graham20183d}, point-based \cite{qi2017pointnet, qi2017pointnet++} and joint fusion methods \cite{su2018splatnet, dai20183dmv, wang2018sgpn}. 
We briefly review these approaches and address the main differences to our work.

\vspace{-4mm}
\paragraph{Image-based} 
Hermans et al.~\cite{hermans2014dense} propose a fast 2D semantic segmentation approach based on randomized decision forests and integrate the semantic segmentation into the 3D reconstruction pipeline. 
McCormac et al.~\cite{mccormac2017semanticfusion} propose a SLAM system that combines CNN to obtain a 3D semantic map. 
However, the above methods generate the predictions purely from 2D images without fully utilizing 3D clues from geometry. 
Our method aims at leveraging all 2D and 3D clues within the scenes.

\vspace{-4mm}
\paragraph{Voxel-based} 
Voxel-based networks consist of a series of 3D convolutional kernels that learn from the input voxelized data \cite{maturana2015voxnet}. 
Dai et al.~\cite{dai2017scannet} propose a voxel-based network for 3D indoor scene semantic segmentation. 
The network takes a sub-volume as input and predicts the class probabilities of the central column. 
However, spatial redundancy occurs in the voxelized data as many voxels remain unoccupied. 
Sparse Convolutional Neural Networks  \cite{graham20183d, graham2015sparse} are proposed to handle the data sparsity by applying kernels on the submanifold area.
They alleviate the computation cost and enable deeper 3D ConvNets with high performance on ScanNet Benchmark \cite{ScanNetBenchmark}. 
However, pre-processes effort for transforming 3D point cloud data into a voxel representation is needed.
The performance of 3D ConvNets rely on voxel resolutions.
Our method directly works on mesh vertices of a 3D scene without the voxelization pre-processing step, which avoids tuning the voxel resolution.

\begin{figure*}[h!]
\centering
\includegraphics[width=\textwidth]{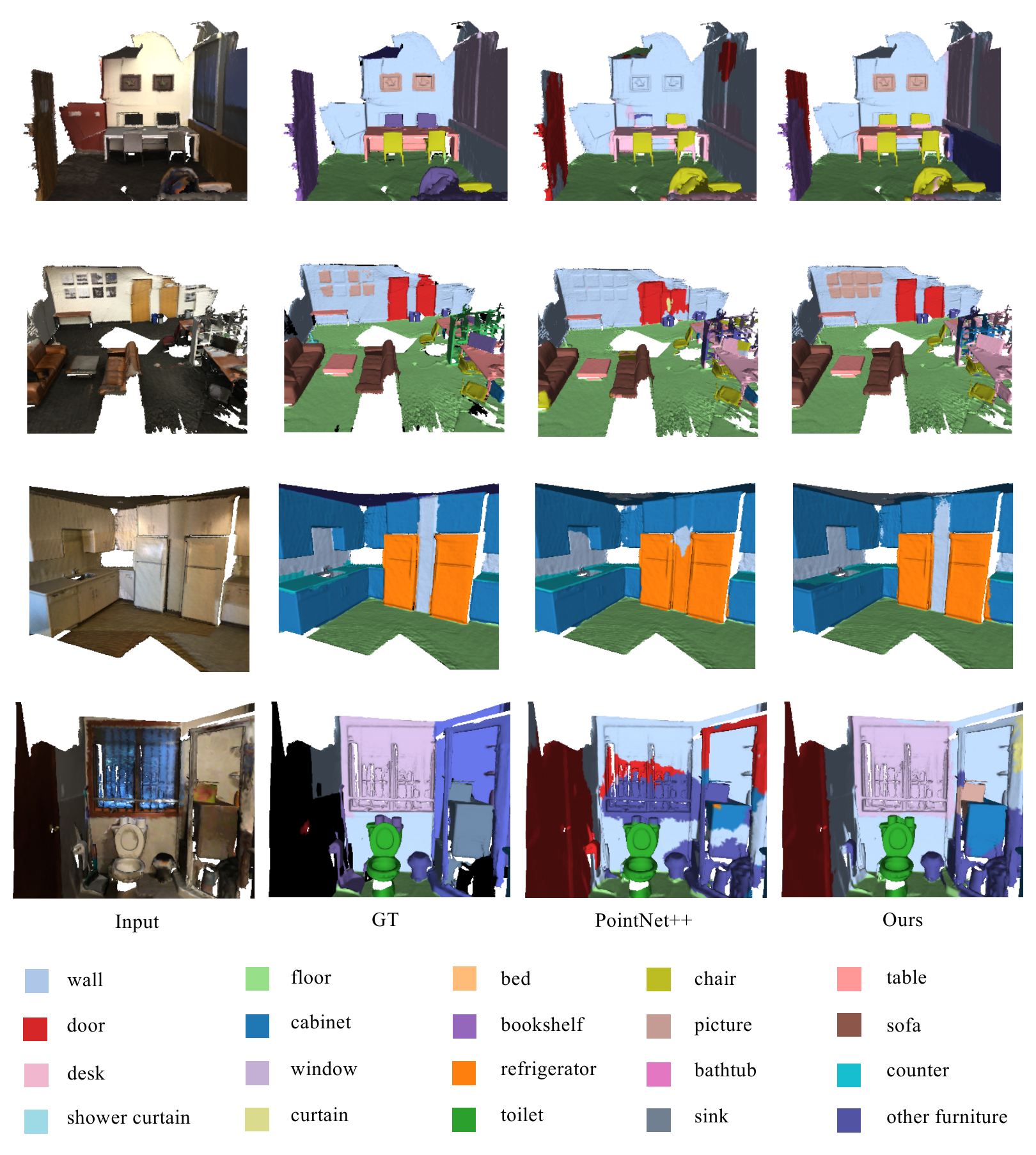}
\caption{
Example prediction results of our system.
We visualize and compare our prediction results with PointNet++ \cite{qi2017pointnet++}. 
Our approach produces better segmentation results and is more capable of recognizing fine-grained details and aware of surrounding context. 
For example, the pictures on the wall in the first row, doors in the second row, the wall between two refrigerators in the third row, and windows and garbage can in the forth row.
} 
\label{fig:result}
\end{figure*}

\vspace{-4mm}
\paragraph{Point-based} 
PointNet \cite{qi2017pointnet} is a pioneer in this direction. 
The authors propose a permutation-invariant network with symmetric function to handle the unordered point sets.
The network basically apply a set of multi-layer perceptron (MLP) networks on each point and aggregate all the point features through a max-pooling layer.
\cite{qi2017pointnet++} improves the PointNet by proposing a hierarchical neural network that captures the fine geometric structures of small neighborhoods.
However, current point-based frameworks do not utilize the information from 2D image features, which are critical for regions lacking explicit structures, such as discriminating a painting from a wall (cf.~Figure \ref{fig:result}).

\vspace{-4mm}
\paragraph{Joint fusion}
Several works address 2D-3D fusion for many tasks.
Liang et al.~\cite{liang2018deep} target on 3D object detection. 
To solve the sparsity of bird's eye view (BEV), they retrieve 2D features as well as 3D features to produce a dense 2D BEV feature map.
Liu et al.~\cite{liu2018floornet} fuse 2D and 3D features for producing semantic 2D floorplans.
However, we focus on developing an algorithm to produce 3D semantic maps.
Dai et al.~\cite{dai20183dmv} extract 2D image features from aligned RGB images and back-project image features into a voxel volume.
Image and 3D geometry streams are jointly fused to predict 3D voxel labels. 
However, their method adopts a volumetric architecture, which lose the input resolution and produce spatial redundancy caused by voxelization.
Su et al.~\cite{su2018splatnet} project 2D and 3D features into a permutohedral lattice and apply sparse convolutions over this sparsely populated lattice. 
They project $n$-dimensional lattice features into a $(n - 1)$-dimensional permutohedral lattice space, which loses one-dimensional structural information.
The sizes of lattice space are controlled by scaling matrices, which also introduce more hyper-parameters and the quantization errors during \textit{splat and slice} steps. 
Wang et al.~\cite{wang2018sgpn} target 3D instance segmentation and extract 2D-3D features from a single 2.5D (RGBD) image and fuse them at the point-level.
However, only partial object surfaces can be observed from a single RGBD.
Compared with the methods aforementioned, our approach is a more generic 3D approach with handling entire 3D scenes and effectively leverages 2D image features, geometric structures and global context priors at point-level within an entire scene.

\section{Proposed Framework} \label{sec:proposed framework}
Given an input 3D mesh and a set of camera poses, which can be estimated using a Structure from Motion system (SfM) or obtained from synthesized camera trajectories, 
the goal of our framework is to produce semantic labels for the input 3D point clouds (i.e., vertices on the 3D mesh). 
Note that our framework is general and applicable to any 3D point clouds and 2D image pairs, not limited to 3D meshes.

Our framework, shown in Figure \ref{fig:network structure}, consists of four parts: 
(1) We apply a 2D CNN to extract appearance features from rendered images and back-project the features into the 3D coordinates. 
The 2D features are interpolated and concatenated with 3D point features as inputs for 3D point-based networks; 
(2) Locally, a sub-volume encoder extracts local fine details in a target 3D sub-volume; 
(3) Globally, global context encoder extracts global scene priors from a sampled sparse scene point sets; 
(4) The decoder aggregates all information: 2D image features, local features and global scene context, and produces the semantic labels for each point in the sub-volume.

\begin{figure}[h!] 
\centering
\includegraphics[width=0.5\textwidth]{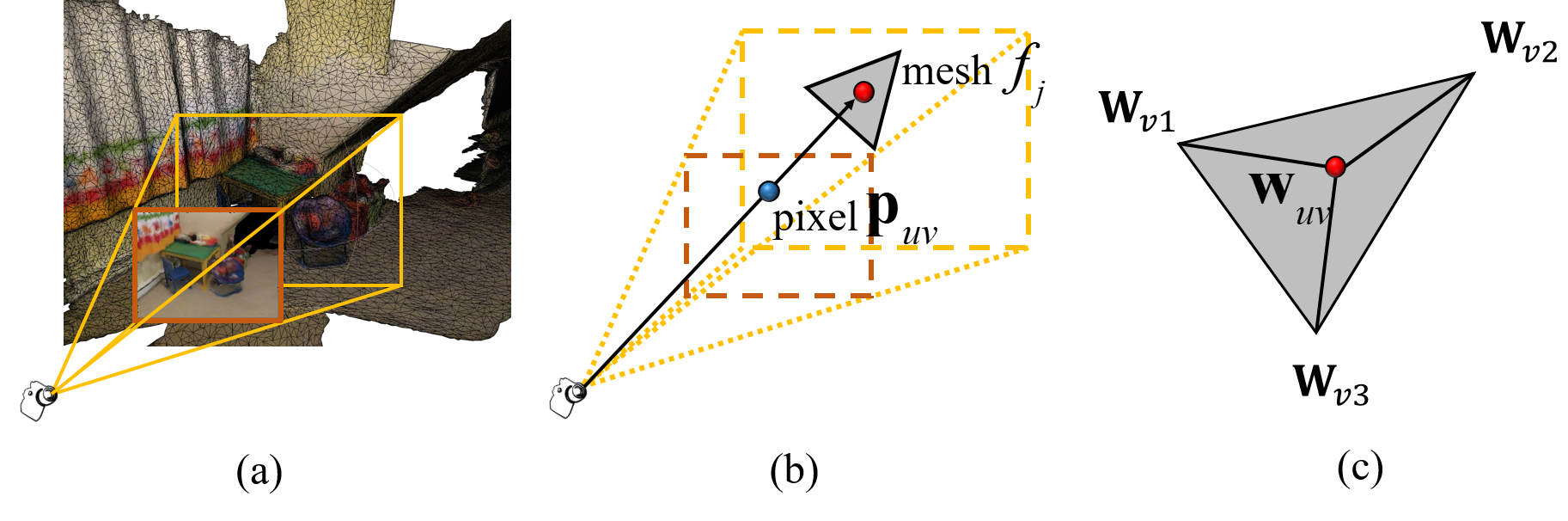}
\caption{(a) We back-project each pixel ${{\bf{p}}_{uv}}$ to 3D space according to the camera intrinsic $[K]$ and extrinsic $[R|T]$ parameters; (b) shows the nearest triangle mesh intersects with the ray direction; (c) We use barycentric interpolation to propagate image features to mesh vertices as the pixels and vertices are not aligned.}
\label{fig:backproject}
\end{figure}

\subsection{Image Features to 3D Vertices}
To obtain the fine-grained appearance features from a complex scene, we render a 3D mesh into 2D images from a series of camera poses, and extract 2D image features by applying a 2D segmentation network. The textural features are back-projected back into 3D spaces by the camera poses.

\subsubsection{Image Feature Extraction}
To extract pixel features from color images, we use DeepLab \cite{chen2018deeplab}, pre-trained on ADE20K \cite{zhou2017scene}, as our 2D segmentation network.
We render color images and ground truth labels from ScanNet 3D meshes to fine-tune DeepLab \cite{chen2018deeplab}. 
We use the layer before the output layer producing a 256-channel feature map with $ (h/4, w/4) $ of the input image size, as our feature descriptors. 
We up-sample the feature map to the original image scale $ (h, w) $ to obtain pixel-level features and back-project the pixelwise 256-dimensional features to 3D mesh vertices. 

\subsubsection{Pixel-vertex Association}
Figure \ref{fig:backproject} illustrates how we associate the 2D image pixels with global 3D points. 
Given the camera extrinsic parameters $[R|T]$ and camera intrinsic parameters $ [K] $, we back-project each pixel from the image coordinate to the 3D world coordinate. 
We calculate the 3D coordinate ${{\bf{W}}_{uv}}$ of each pixel ${{\bf{p}}_{uv}}$.
We associate the pixel with the triangle mesh $ f_{j} $ where ${{\bf{W}}_{uv}}$ lies.
The world coordinates of three vertices $({{\bf{W}}_{{v_1}}},{{\bf{W}}_{{v_2}}},{{\bf{W}}_{{v_3}}})$ that compose the triangle mesh $ f_{j} $ are used to compute barycentric weights.

\subsubsection{Barycentric Interpolation}
Since each pixel finds the three corresponding vertices of the triangle mesh that  $\bf{{W}_{uv}} $ lies in, we propagate the image features to three triangle vertices via barycentric interpolation.
For each vertex, we sum 256-dimensional features from all pixels and normalize the feature to a unit-length vector.
For those vertices that are occluded by 3D geometry and do not have any mapped 2D image features, we fill zero vectors for the vertices.
\subsection{Geometry Structure}
\label{subsec:geometry_structure}
3D scenes have varied layouts and contain multiple objects with different poses and locations.
To encode a complex scene, we propose two point-based architectures that learn the structure details in a sub-volume and the global context priors respectively.
The decoder predicts semantic labels for the points in the sub-volume from aggregated features.  

\subsubsection{Sub-volume Encoder}
For each sub-volume, we extract a dense point set in order to preserve the structure details. 
In our experiments, we sample 8192 points for each sub-volume, approximately $50\%$ of the points in a sub-volume.  
Besides the Euclidean coordinates $(x, y, z)$, we concatenate 256-dimensional features from DeepLab \cite{chen2018deeplab} and the vertex normal $({n_x},{n_y},{n_z})$ resulting in a 262-dimensional feature $(x,y,z,{n_x},{n_y},{n_z},{\bf{d}})$ of a point.
We apply a series of sampling, grouping and multi-layer perceptron (MLP) as in \cite{qi2017pointnet++} to reduce the number of points and extract features to represent each point in the sub-volume. 
We use four layers to encode the input point set and the number of output points of each layer are 1024, 256, 64, 16 respectively.
As our input feature dimension is much larger than the original PointNet++ \cite{qi2017pointnet++} (262 vs. 3), we increase the dimensions of each MLP layer. 
The parameters for each layer are [1024, 0.1, 32, (128, 128)], [256, 0.2, 32, (256, 256)], [64, 0.4, 32, (512, 512)], [16, 0.8, 32, (512, 512, 726)], where the parameters are in the format of [number of sample points, grouping radius (in meter), number of points in a group, (MLP output dimensions)].
Note that each layer consists of multiple MLPs.
%
%
%

\subsubsection{Global Scene Encoder}
We introduce a global scene encoder to learn the global priors from a relatively sparse point set that covers the whole scene. 
In our experiments, we sample 16384 points, which is approximately 10\% of mesh vertices from the entire scene.
Each point is associated with 262-dimensional features as in the sub-volume encoder.
The sparse scene points with 2D image features are fed into the global context encoder to obtain the global context features. 
Similar to the sub-volume encoder, four layers are used. 
The number of sample points of each layer are 4096, 1024, 256, 128 respectively.
The parameters in each layer are [(4096, 0.4, 32, (128, 128)], [1024, 0.8, 32, (256, 256)], [256, 1.2, 32, (512, 512)], [128, 1.6, 32, (512, 512, 726)].

\subsubsection{Decoder}
The decoder network consisting of four layers is proposed to learn the textural, local geometry and global priors features encoded by the encoders.
A layer in the decoder consists of three parts: feature concatenation, multi-layer perceptron (MLP) and feature propagation.

\vspace{-4mm}
\paragraph{Feature concatenation} 
We concatenate three different features for each point in the target sub-volume. 
(1) The interpolated features from the previous layer (blue color in Figure \ref{fig:network structure});
(2) the interpolated features from the global scene encoder (orange color in Figure \ref{fig:network structure}); 
(3) the features from the skip layers in the sub-volume encoder (gray color in Figure \ref{fig:network structure}).

\vspace{-4mm}
\paragraph{MLP and feature propagation}
Similar to the encoder, each layer consists of multiple MLPs.
In our experiment, we use two MLP layers with 256 dimensions in each layer. 
We apply feature propagation layers as in \cite{qi2017pointnet++} to upsample the points from the previous layer.

Four layers are deployed to decode the aggregated features and upsample points from bottleneck size to original input size, which are $(16, 64, 256, 1024, input\_size)$.
Finally, the point features pass through an output layer consisting of two MLPs to produce the class probabilities for each point in the sub-volume.

\subsection{Overlapping Sub-volumes}
\label{subsec:overlapping}
Since our network produces the points' class probabilities in a sub-volume, we utilize a sliding window strategy to obtain the final points' predictions in a whole scene. 
As shown in Table \ref{tb:stride size}, the performance is significantly improved by overlapping sliding windows. 
We sum the class probability of each point in the overlapping region and select the class with the maximum score. 
In our experiments, we use the window size $ height \times width \times depth = {\textit{scene height}} \times 1.5m \times 1.5m$ and stride size $0.45m$, which produces good results in a reasonable time, please see Table \ref{tb:stride size} for more details. 

\begin{table*}[!t]
\resizebox{\textwidth}{!}{\begin{tabular}{l | |*{19}{c}r|c}
Method 							& floor 	& wall 	& chair 	& sofa 	& table 	& door 	& cab   	& bed 	& desk  	& toil 	& sink    	& wind   	& pic 	& bkshf   	& curt    	& show 	& cntr    	& fridg 	& bath   	& other   	& mIOU   \\  \hline
DeepLab \cite{chen2018deeplab}		& 82.4  	& 70.9 	& 61.6  	& 59.2 	& 52.5  	& 48.8 	& 48.1  	& 67.3 	& 47.2  	& 72.9 	& 53.9   	& 55.1    	& 33.6 	& 60.9     	& 54.7   	& 45.5 	& 52.8   	& 38.2 	& 66.5   	& 38.4     	& 55.5 	\\ 
PointNet++ \cite{qi2017pointnet++}   		& 93.7  	& 72.2 	& 80.7  	& 66.0 	& 62.1  	& 34.1 	& 43.4  	& 66.2 	& 48 	    	& 81.7 	& 54.3   	& 39.6 	& 8.1 	& 67.1     	& 30.8   	& 34.2 	& 50.2   	& 30.9 	& 73.9   	& 34.3     	& 53.5 	\\ 
3D Sparse Conv \footnotemark[1] \cite{graham20183d}		        		& 93.9   	& 77.4   	& 82.4   	& 73.0   	& 64.2  	& 45.7  	& 54.3  	& 76.4  	& 54.8  	& 75.2  	& 50.7   	& 50.1   	& 22.8    	&71.5 & 53.9     	& 50.7     	& 55.6      	&37.0 & 76.4    	& 44.7     	& 60.5      \\ 
\hline\hline
\bf{Ours w/ SfM poses}   			& 95.4 & 82.4 & 86.9 & \bf{73.0} & \bf{71.2} & 58.4 & 57.1 & 80.5 & 60.7 & \bf{90.8} & \bf{60.7} & 62.8 & 35.6 & 77.3 & 68.5 & 63.1 & 59.7 & \bf{50.3} & \bf{83.7} & 55.8 & 68.2  \\
\bf{Ours w/ syn poses}   		    & \bf{95.5} & \bf{84.0} & \bf{87.9} & 72.5 & 71.1 & \bf{60.1} & \bf{65.5} & \bf{80.9} & \bf{61.0} & 87.0 & 60.4 & \bf{63.2} & \bf{39.2} & \bf{78.0} & \bf{72.4} & \bf{64.4} & \bf{60.1} & 46.1 & 77.5 & \bf{57.4} & \bf{69.2}
\end{tabular}}
\caption{
Results on ScanNet validation set. 
Our framework shows significantly better mIoU scores than DeepLab \cite{chen2018deeplab}, PointNet++ \cite{qi2017pointnet++} and 3D Sparse Conv \cite{graham20183d} for 3D semantic segmentation.}
\label{tb:validation set}
\end{table*}

\begin{table*}[!t]
\resizebox{\textwidth}{!}{\begin{tabular}{l | |*{19}{c}r|c}
Method 							& floor 	& wall 	& chair  	& sofa 	& table  	& door 	& cab   	& bed 	& desk  	& toil   	& sink    	& wind    	& pic       	& bkshf   	& curt    	& show   	& cntr     	& fridg   	& bath   	& other    	& mIOU    \\  \hline
ScanNet \cite{dai2017scannet}			& 78.6  	& 43.7  	& 52.4   	& 34.8   	& 30     	& 18.9  	& 31.1  	& 36.6   	& 34.2  	& 46    	& 31.8   	& 18.2    	& 10.2    	& 50.1    	& 0.2      	& 15.2     	&  21.1  	&  24.5  	& 20.3   	& 14.5     	& 30.6      \\ 
PointNet++ \cite{qi2017pointnet++}   		& 78.6  	& 52.3  	& 36      	& 34.6   	& 23.2  	& 261   	& 25.6  	& 47.8   	& 27.8  	& 54.8 	& 36.4   	& 25.2    	& 11.7    	& 45.8    	& 24.7    	& 14.5     	&  25      	&  21.2  	& 58.4   	& 18.2    	& 33.9       \\ 
SplatNet \cite{su2018splatnet}        	 	& 92.7  	& 69.9   	& 65.6   	& 51      	& 38.3  	& 19.7  	& 31.1  	& 51.1  	& 32.8  	& 59.3  	& 27.1   	& 26.7    	& 0         	& 60.6   	& 40.5     	& 24.9     	& 24.5   	& 0.1    	& 47.2    	& 22.7     	& 39.3      \\
3DMV \cite{dai20183dmv}		        		& 79.6   	& 53.9   	& 60.6   	& 50.7   	& 41.3  	& 37.8  	& 42.4  	& 53.8  	& 43.3  	& 69.3  	& 47.2   	& 53.9   	& 21.4    	&64.3 & 57.4     	& 20.8     	& 31      	&53.7 & 48.4    	& 30.1     	& 48.4      \\ 
3D Sparse Conv\footnotemark[1] \cite{graham20183d}          & 93.7   	& 79.9   	& 76.2   	& 66.3   	& 49.6  	& 45.2  	& 62.0  	& 75.0  	& 46.9  	& 80.8  	& 54.0   	& 58.6   	& 27.5    	&\bf{68.8} & 56.1     	& \bf{53.7}     	& \bf{43.3}      	&54.0 & 59.8    	& 44.1     	& 59.8      \\  

\hline\hline

\bf{Ours w/ SfM poses}           	        & 94.3 	& 79.5 & 79.5 & 74.4 	&\bf{57} 	&53.9 	&57.1 	&74.6 	&\bf{48.5} 	&\bf{85.9} 	&\bf{63.5} &\bf{62.8} 	&28.7 	& 61.2 	&79.8 	&41.8 	&38.6 	& 52 		&\bf{64.5} 	&44.5 	&62.1	\\ 

\bf{Ours w/ syn poses}           	        &\bf{95.1} 	&\bf{81.4} &\bf{82.5} &\bf{76.4} 	&55.9 	&\bf{56.1} 	&\bf{63.3} 	&\bf{77.8} 	&46.7 	&83.8 	&57.9 &59.8 	&\bf{29.1} 	&66.7 	&\bf{80.4} 	&45.8 	&42.0 	& \bf{56.6} 		& 61.4 	&\bf{49.4} 	&\bf{63.4}	\\  

\end{tabular}}
\caption{
Results on ScanNet testing set.
Our unified framework outperforms several state-of-the-art methods by a large margin. 
The gain comes from jointly optimizing 2D image features, 3D structures and global context in a point-based architecture. 
With synthetic camera poses, we can further improve our performance ($+1.3\%$ gains)}  
\label{tb:test set}
\end{table*}

\section{Experiments}  \label{sec:experiment}
\paragraph{ScanNet Benchmark \cite{ScanNetBenchmark}}
This dataset is proposed in 2017 by Dai et al.~\cite{dai2017scannet} for 2D and 3D indoor scene semantic segmentation, and is currently the largest and most challenging RGB-D reconstruction dataset.
It contains 1513 RGB-D indoor scans and provides with both 3D vertex labels and 2D dense pixel labels, as well as corresponding camera parameters. 
We use the train/validation split provided by ScanNet, 1201 for training and 312 for validation. 
We perform the 20-class semantic segmentation task defined in the benchmark.

\vspace{-4mm}
\paragraph{Evaluation metrics}
We follow the 3D evaluation metrics in ScanNet benchmark \cite{ScanNetBenchmark}. 
It computes the Mean Intersect over Union (mIoU) score between the predicted labels and the ground truth labels. 
We use the evaluation script provided by ScanNet benchmark to obtain our validation scores, and upload our results to the online evaluation system for our testing scores.

\vspace{-4mm}
\paragraph{Training details}
For each scene, we randomly sample 8192 points for a sub-volume with the size of $height \times width \times depth = 3m \times 1.5m \times 1.5m$ and 16384 points for the entire scene as our training samples. 
We check the label distribution of each sub-volume and discard the sub-volume with less than $70\%$ annotated vertices. 
We randomly rotate the entire scene points along the z-axis for data augmentation.
We set the batch size to $6$ on one Nvidia P100 GPU and deploy our model on two GPUs during training. 
The optimizer is Adam Optimizer. 
The initial learning rate is $0.001$ and decays every $2000$ steps with a power of $0.9$. 
Weighted cross entropy loss is adopted to deal with unbalanced ground truth classes.

\vspace{-4mm}
\paragraph{Testing details}
During testing time, we predict each point's class label within the sub-volume region. 
We slide the sub-volume and overlap the prediction region to produce an entire scene semantic map and to improve the performance.
We pad $0.5m$ along the X-Y axis and slide the sub-volume through the entire scene with the window size ${\textit{scene height}} \times 1.5m \times 1.5m$.
We set the stride size to $0.45m$ in both X and Y direction. 

\vspace{-4mm}
\paragraph{Baseline methods}
We compare our framework with several state-of-the-art methods, including DeepLab \cite{chen2018deeplab},  ScanNet \cite{dai2017scannet}, PointNet++ \cite{qi2017pointnet++}, 3DMV \cite{dai20183dmv}, SparseConv \cite{graham20183d} and SplatNet \cite{su2018splatnet} on ScanNet validation and testing set. 
For validation set (cf.~Table \ref{tb:validation set}), we use the original source codes from DeepLab \cite{chen2018deeplab}, PointNet++ \cite{qi2017pointnet++}, 3D Sparse Conv \cite{graham20183d} and fine-tune their model on ScanNet training set.
For DeepLab \cite{chen2018deeplab}, we predict one frame sampled from every 20 frames using the same sampling rate as our method.
We back-project the class probability of each pixel into the 3D coordinates and compute the class probability of each vertex via barycentric interpolation. 
We aggregate the class probabilities from all sampled frames to obtain the final prediction.
For PointNet++ \cite{qi2017pointnet++}, we also perform the overlapping sliding window and use the same window size (${\textit{scene height}}  \times 1.5 m \times 1.5 m$) as our settings for fairly comparison. 
We compare 3D Sparse Conv \cite{graham20183d} as one of our baselines on validation set with $5 cm^3$ color voxels and using 3D UNet \cite{cciccek20163d} architecture.
Note that we use $5 cm^3$ color voxels in 3D Sparse Conv \cite{graham20183d} experiment in order to fairly compare with 3DMV \cite{dai20183dmv}, though 3D Sparse Conv reaches higher performance with $2 cm^3$ voxels when using 3D ResNet \cite{he2016deep} as the backbone network.
For testing set (cf.~Table \ref{tb:test set}), we compare our method with the reported scores of ScanNet \cite{dai2017scannet}, PointNet++ \cite{qi2017pointnet++}, 3DMV \cite{dai20183dmv} and SplatNet \cite{su2018splatnet} from the ScanNet benchmark \cite{ScanNetBenchmark} leader board.

\footnotetext[1]{We use $5 cm^3$ color voxels in 3D Sparse Conv \cite{graham20183d} experiment for fairly comparing with 3DMV \cite{dai20183dmv}.}

\begin{table*}[!t]
\resizebox{\textwidth}{!}{\begin{tabular}{l | |*{19}{c}r|c}
Features						            & floor & wall  & chair & sofa  & table & door  & cab   & bed   & desk  & toil  & sink  & wind  & pic   & bkshf & curt  & show  & cntr  & fridg & bath  & other   & mIOU  \\  \hline
$xyz$                         	            & 93.7  & 72.2  & 80.7  & 66.0  & 62.1  & 34.1  & 43.4  & 66.2  & 48.0  & 81.7  & 54.3  & 39.6  & 8.1   & 67.1  & 30.8  & 34.2  & 50.2  & 30.9  & 73.9    & 34.3  & 53.5  \\
$xyz$ + $gc$                                    & 93.8  & 72.0  & 80.8  & 72.9  & 60.5  & 39.6  & 48.3    & 69.9  & 51.9  & 74.0  & 54.0  & 41.4  & 9.8   & 63.1  & 53.7  & 50.9  & 55.0  & 40.2  & 77.7  & 33.4  & 57.1  \\
$xyz$ + ${\bf{n}}$                            & 94.0  & 75.4  & 83.9  & 69.5  & 65.2  & 38.1  & 51.4    & 68.4  & 51.7  & 82.9  & 55.9  & 47.5  & 17.3  & 73.3  & 59.2  & 42.1  & 58.7  & 42.1  & 80.3  & 42.3  & 60.0  \\
$xyz$ + ${\bf{n}}$ + $gc$         	            & 94.9  & 76.5  & 84.2  & \bf{77.0}  & 64.7  & 43.9  & 55.5    & 75.5  & 55.8  & 82.7  & 57.1  & 46.8  & 18.8  & 72.9  & 61.1  & 51.1  & \bf{59.8}  & 44.1  & 82.4  & 40.6  & 62.2  \\ 
$xyz$ + ${\bf{n}}$ + ${\bf{d}}$     		& 94.6  	& 80.7 	& 85.2  	& 71.4  	& 69   & 56.3  	& 55.3  	& 79.1   	& 57.9  	& 86.2  	& \bf{60.9}   	& 59.8   	& 34.4     	& 75.9     	& 66.4   	& 58.4  	& 59.3 & 49.1  	& 76.3   	& 50.7     	& 66.3       \\	
$xyz$ + ${\bf{n}}$ + ${\bf{d}}$ + $gc$     & \bf{95.4}  & \bf{82.4}  & \bf{86.9}  & 73.0  & \bf{71.2}  & \bf{58.4}  & \bf{57.1}    & \bf{80.5}  & \bf{60.7}  & \bf{90.8}  & 60.7  & \bf{62.8}  & \bf{35.6}  & \bf{77.3}  & \bf{68.5}  & \bf{63.1}  & 59.7  & \bf{50.3}  & \bf{83.7}  & \bf{55.8}  & \bf{68.2}  \\ 										  
\end{tabular}}
\caption{Different features of our unified framework are analyzed.
The results show that 3D points ($xyz$), vertex normal (${\bf{n}}$), global context ($gc$) and 2D image features (${\bf{d}}$) benefit to each other.
They lead to the improved performance of our framework.} 
\label{tb:features}
\end{table*}

\section{3D Point Cloud Segmentation Results} \label{sec:segmentation results}
Table \ref{tb:validation set} and Table \ref{tb:test set} summarize the average IOU for different methods on the ScanNet validation and testing set respectively.  
Our unified model outperforms the existing fusion-based methods, 3DMV \cite{dai20183dmv}, SplatNet \cite{su2018splatnet} on the testing set.
We conclude the reasons as follow: 
(1) Our framework uniformly samples the points on the object surfaces while voxel-based approaches divide the input space to voxels, which has quantization errors;
(2) We simultaneously optimize the 2D textural, 3D geometrical and global context feature within a point-based framework for better predictions;
(3) We preserve all the structure information without projecting the high-dimensional points to a hyper-plane as in  \cite{su2018splatnet}.
%

\section{Ablation Studies} \label{sec:ablation studies}
We provide an in-depth analysis of features, stride size of the sliding window, and synthetic camera pose in the following paragraphs.

\subsection{Feature Analysis}
We evaluate different types of features in our unified framework: 3D coordinates ($xyz$), vertex normal (${\bf{n}}$), global context ($gc$) and 2D image features (${\bf{d}}$), as shown in Table \ref{tb:features}. 
Noted that the first row ($xyz$) in Table \ref{tb:features} is equal to PointNet++ \cite{qi2017pointnet++} in Table \ref{tb:validation set};
We also add normal vectors to each point in PointNet++ \cite{qi2017pointnet++} setting, which is the third row ($xyz$ + $\bf{n}$) in Table \ref{tb:features}.

\vspace{-4mm}
\paragraph{Global contexts} improve overall performance in all cases: $53.5\%$ ($xyz$) vs. $57.1\%$ ($xyz$ + $gc$), $60.0\%$ ($xyz$ + $\bf{n}$) vs. $62.2\%$ ($xyz$ + $\bf{n}$ + $gc$), $66.3\%$ ($xyz$ + $\bf{n}$ + $\bf{d}$) vs. $68.2\%$ ($xyz$ + $\bf{n}$ + $\bf{d}$ + $gc$).
Note that global priors are not limit to single room cases, it generalizes to 21 scene types in ScanNet benchmark. 
We observe an interesting example that curtain and shower curtain are very difficult to distinguish without knowing the scene context information.
By incorporating the global prior, the accuracy of both classes are significantly increased  (curtain from 66.4\% to 69.6\% and shower curtain from 58.4\% to 64.3\%).

\vspace{-4mm}
\paragraph{Vertex normal} improves our framework from $57.1\%$ ($xyz$ + $gc$) to $62.2\%$ ($xyz$ + $\bf{n}$ + $gc$), suggesting that normal vectors effectively describe the 3D scene and object structures. For example, bathtub, chair and table have very different geometric structures.

\vspace{-4mm}
\paragraph{Image features} also show significant improvement to the overall performance, as 2D CNN is pre-trained on a large-scale image dataset and fine-tuned on high-resolution 2D images, which helps discriminate fine-grained details of objects without explicit structures, such as a painting on a wall. As a result, our method leads to better performance comparing to color voxel used in 3D Sparse Conv (cf.~Table \ref{tb:validation set} and Table \ref{tb:test set}).

In conclusion, we demonstrate that normal vectors, color features and global priors contain different semantics in our experiments shown in Table \ref{tb:features}.
We improve the performances as the additional information is fused into our framework: $53.5\%$ ($xyz$), $60.0\%$ ($xyz$ + ${\bf{n}}$), $62.2\%$ ($xyz$ + ${\bf{n}}$ + $gc$), $68.2\%$ ($xyz$ + ${\bf{n}}$ + ${\bf{d}}$ + $gc$).
It proves that texture, geometry and global context encodes different information for semantics respectively.

\begin{table}[ht]
\centering
\begin{tabular}{|c|c|c|c|c|}
\hline
Stride size  	&  1.5 m   	& 1.05m 	 & 0.75 m  & 0.45 m  \\ \hline
mIoU         	&  59.5 \% & 60.8 \% & 61.8 \% & 62.2 \% \\ \hline
\end{tabular}
\caption{Analysis of sliding sub-volume stride sizes. The results show that overlapping slide window improves the performance.}
\label{tb:stride size}
\end{table}

\subsection{Sub-volume Stride}
In Table \ref{tb:stride size}, we evaluate different stride sizes ranging from $0.45 m$ to $1.5 m$, where a stride size of $1.5 m$ means no overlapping as our sub-volume size is set to ${\textit{scene height}} \times 1.5 m \times 1.5 m$. 
With stride size $=1.05 m$ in both depth and width yields 1.3\% improvement in mIoU score.
The performance is further boosted to 62.2 \% (2.7 \% gains) for stride size $=0.45 m$. 
We only see marginal improvement for smaller stride sizes, but the time complexity will be exponentially increased. 
Therefore, we set our stride size to $0.45 m$ shows good performance while it runs in a reasonable time. 

%

\begin{table}[!t]
\footnotesize
\centering
\begin{tabular}{l||c|c|c}
Number of images    & Vertex coverage   & mIOU      \\ \hline
61447               & 98.8\%            &  69.2\%   \\ 
30723               & 98.6\%            &  68.7\%   \\ 
18433               & 98.3\%            &  68.1\%   \\ 
6144                & 95.2\%            &  66.8\%   \\ 
\end{tabular}
\caption{
We investigate different numbers of rendered images by using the synthetic camera poses.}
\label{tb:synthetic camera pose}
\end{table}

\begin{figure}[!htbp]
\centering
\includegraphics[width=0.5\textwidth]{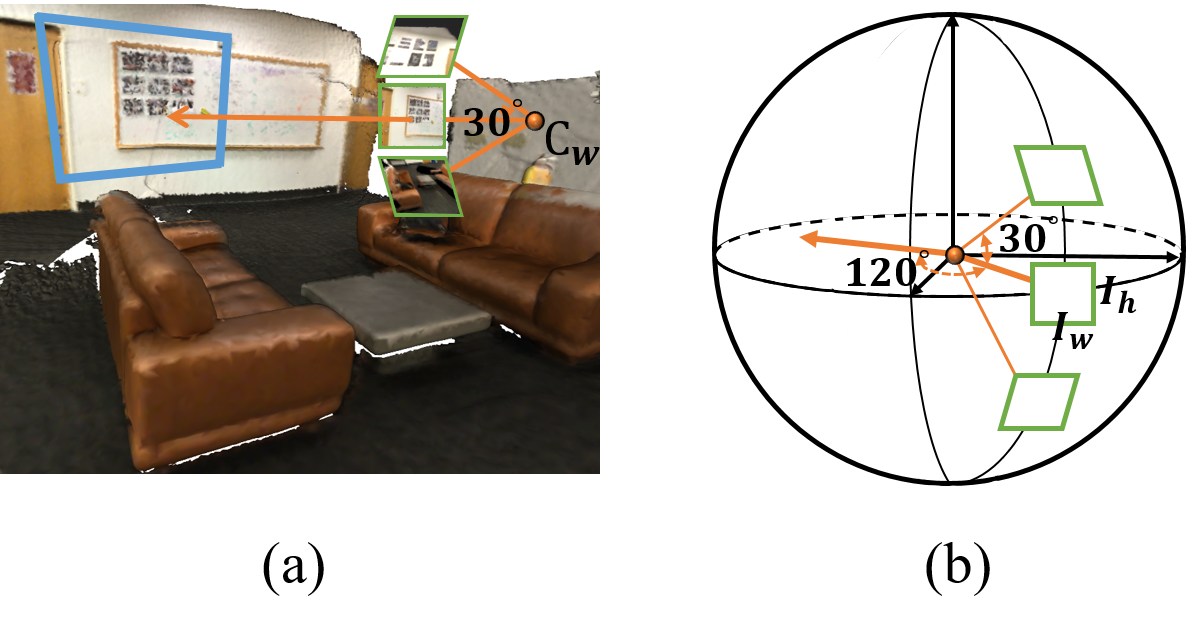}
\caption{
(a) We put our synthetic camera model at a position $C_W = (h_{c}, w_{c}, d_{c})$ in the scene.
(b) For each selected position, we define three attitude degrees, which are -30\degree, 0\degree, 30\degree and rotate our camera 120\degree in azimuth.
} 
\label{fig:synthetic camera}
\end{figure}

\begin{figure*}[h!]
\centering
\includegraphics[width=\textwidth]{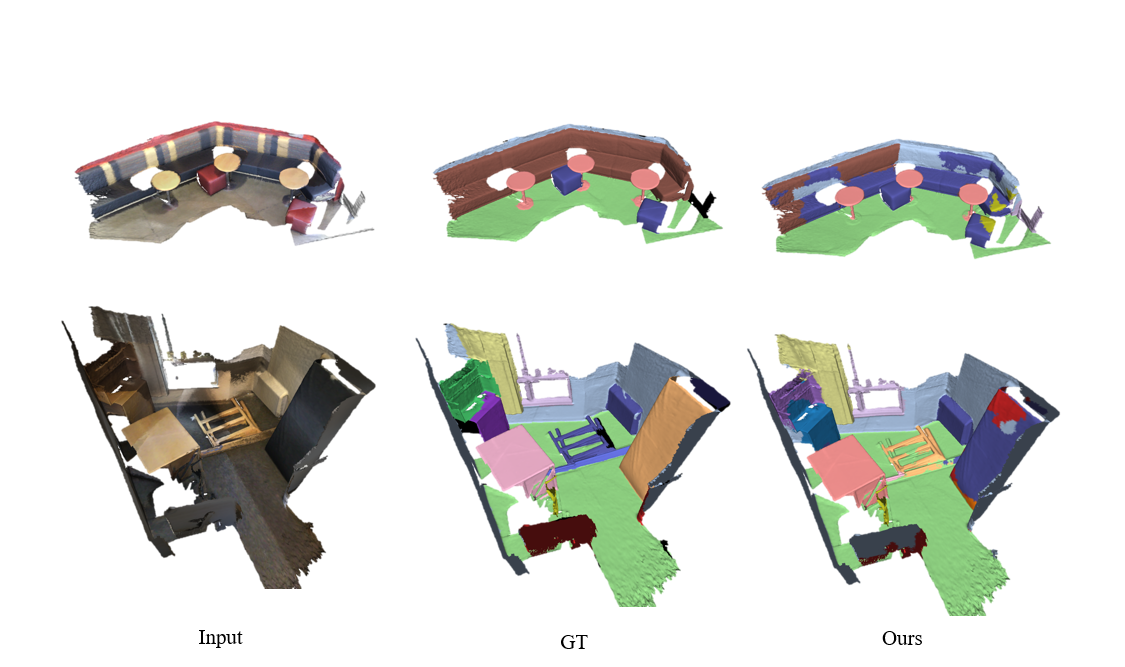}
\caption{Some failure cases of our system.  
Top row shows that our system fails to predict the circular sofa due to the unusual shape that is very different from our training examples.
Bottom row shows some failure cases: (1) fine-grained objects. e.g., desk (pink color) is mis-classified as table (orchid color), (2) large color and shape variations for ``other furniture'' class, e.g., two blue objects in the middle of the scene are labeled as other furniture class, and (3) unusual pose, e.g., an upright bed.}
\label{fig:failure}
\end{figure*}

\subsection{Synthetic Camera Model}
\label{subsec:synthetic_camera}
We explore synthetic camera model to improve the wrongly estimated camera pose from SfM, which affects the 3D segmentation performance in ScanNet benchmark.  
We slice each scene into three levels of heights: $1.5m$, $2m$, and $2.5m$.
At each level, we equally divide the scene width and scene depth into $10$ partitions ($scene\,width/10, scene\,depth/10$).
As shown in Figure \ref{fig:synthetic camera} (a) and (b), for each camera position, we render $9$ images (attitude: $-30\degree$, $0\degree$, $30\degree$; azimuth: $0\degree$, $120\degree$, $240\degree$).
As a result, we capture $2700$ ($10\times10\times3\times9$) images with resolution $(I_h, I_w) = (480, 640)$ for each scene.
We discard the images with insufficient context and select the image set that has the highest coverage of the scene vertices, resulting in $61446$ and $23146$ images in total for the validation set and the testing set.
Table \ref{tb:synthetic camera pose} shows the experiments of using different numbers of rendered images from synthetic camera poses, which improves the segmentation results from structure from motion (i.e., 68.2\%).  
%
%
%
%

\section{Conclusions}
We have presented a unified point-based framework for optimizing 2D image features, 3D structures and global context priors. 
By leveraging global context priors, we improved 3D semantic segmentation performance over several state-of-the-art methods in the ScanNet benchmark \cite{ScanNetBenchmark}, confirming the ability of our model to deliver more informative features than previous work. 
Our in-depth feature analysis proves that textures, geometry and global context encode different meanings for semantics.
We also showed that overlapping sub-volume and synthetic camera poses further improve the prediction results.

\section*{Acknowledgement}
This work was supported in part by the Ministry of Science and Technology, Taiwan, under Grant MOST 108-2634-F-002-004 and FIH Mobile Limited. We also benefit from the NVIDIA grants and the DGX-1 AI Supercomputer.

\balance

\clearpage

{\small
\bibliographystyle{ieee}
\bibliography{egbib}

\begin{thebibliography}{10}\itemsep=-1pt

\bibitem{chang2017matterport3d}
A.~Chang, A.~Dai, T.~Funkhouser, M.~Halber, M.~Nie{\ss}ner, M.~Savva, S.~Song,
  A.~Zeng, and Y.~Zhang.
\newblock Matterport3d: Learning from rgb-d data in indoor environments.
\newblock {\em arXiv preprint arXiv:1709.06158}, 2017.

\bibitem{chen2018deeplab}
L.-C. Chen, G.~Papandreou, I.~Kokkinos, K.~Murphy, and A.~L. Yuille.
\newblock Deeplab: Semantic image segmentation with deep convolutional nets,
  atrous convolution, and fully connected crfs.
\newblock {\em TPAMI}, 2018.

\bibitem{cciccek20163d}
{\"O}.~{\c{C}}i{\c{c}}ek, A.~Abdulkadir, S.~S. Lienkamp, T.~Brox, and
  O.~Ronneberger.
\newblock 3d u-net: learning dense volumetric segmentation from sparse
  annotation.
\newblock In {\em MICCAI}, 2016.

\bibitem{ScanNetBenchmark}
A.~Dai, A.~X. Chang, M.~Savva, M.~Halber, T.~Funkhouser, and M.~Nie{\ss}ner.
\newblock \url{http://kaldir.vc.in.tum.de/scannet_benchmark/}.

\bibitem{dai2017scannet}
A.~Dai, A.~X. Chang, M.~Savva, M.~Halber, T.~A. Funkhouser, and M.~Nie{\ss}ner.
\newblock Scannet: Richly-annotated 3d reconstructions of indoor scenes.
\newblock In {\em CVPR}, 2017.

\bibitem{dai20183dmv}
A.~Dai and M.~Nie{\ss}ner.
\newblock 3dmv: Joint 3d-multi-view prediction for 3d semantic scene
  segmentation.
\newblock In {\em ECCV}, 2018.

\bibitem{graham2015sparse}
B.~Graham.
\newblock Sparse 3d convolutional neural networks.
\newblock In {\em BMVC}, 2015.

\bibitem{graham20183d}
B.~Graham, M.~Engelcke, and L.~van~der Maaten.
\newblock 3d semantic segmentation with submanifold sparse convolutional
  networks.
\newblock In {\em CVPR}, 2018.

\bibitem{he2016deep}
K.~He, X.~Zhang, S.~Ren, and J.~Sun.
\newblock Deep residual learning for image recognition.
\newblock In {\em CVPR}, 2016.

\bibitem{hermans2014dense}
A.~Hermans, G.~Floros, and B.~Leibe.
\newblock Dense 3d semantic mapping of indoor scenes from rgb-d images.
\newblock In {\em ICRA}, 2014.

\bibitem{liang2018deep}
M.~Liang, B.~Yang, S.~Wang, and R.~Urtasun.
\newblock Deep continuous fusion for multi-sensor 3d object detection.
\newblock In {\em ECCV}, 2018.

\bibitem{liu2018floornet}
C.~Liu, J.~Wu, and Y.~Furukawa.
\newblock Floornet: A unified framework for floorplan reconstruction from 3d
  scans.
\newblock In {\em ECCV}, 2018.

\bibitem{ParseNet}
W.~Liu, A.~Rabinovich, and A.~C. Berg.
\newblock Parsenet: Looking wider to see better.
\newblock In {\em ICLR}, 2016.

\bibitem{maturana2015voxnet}
D.~Maturana and S.~Scherer.
\newblock Voxnet: A 3d convolutional neural network for real-time object
  recognition.
\newblock In {\em IROS}, 2015.

\bibitem{mccormac2017semanticfusion}
J.~McCormac, A.~Handa, A.~Davison, and S.~Leutenegger.
\newblock Semanticfusion: Dense 3d semantic mapping with convolutional neural
  networks.
\newblock In {\em ICRA}, 2017.

\bibitem{qi2017pointnet}
C.~R. Qi, H.~Su, K.~Mo, and L.~J. Guibas.
\newblock Pointnet: Deep learning on point sets for 3d classification and
  segmentation.
\newblock In {\em CVPR}, 2017.

\bibitem{qi2017pointnet++}
C.~R. Qi, L.~Yi, H.~Su, and L.~J. Guibas.
\newblock Pointnet++: Deep hierarchical feature learning on point sets in a
  metric space.
\newblock In {\em NIPS}, 2017.

\bibitem{su2018splatnet}
H.~Su, V.~Jampani, D.~Sun, S.~Maji, E.~Kalogerakis, M.-H. Yang, and J.~Kautz.
\newblock Splatnet: Sparse lattice networks for point cloud processing.
\newblock In {\em CVPR}, 2018.

\bibitem{su2015multi}
H.~Su, S.~Maji, E.~Kalogerakis, and E.~Learned-Miller.
\newblock Multi-view convolutional neural networks for 3d shape recognition.
\newblock In {\em ICCV}, 2015.

\bibitem{tchapmi2017segcloud}
L.~Tchapmi, C.~Choy, I.~Armeni, J.~Gwak, and S.~Savarese.
\newblock Segcloud: Semantic segmentation of 3d point clouds.
\newblock In {\em 3DV}, 2017.

\bibitem{wang2018sgpn}
W.~Wang, R.~Yu, Q.~Huang, and U.~Neumann.
\newblock Sgpn: Similarity group proposal network for 3d point cloud instance
  segmentation.
\newblock In {\em CVPR}, 2018.

\bibitem{zhao2017pyramid}
H.~Zhao, J.~Shi, X.~Qi, X.~Wang, and J.~Jia.
\newblock Pyramid scene parsing network.
\newblock In {\em CVPR}, 2017.

\bibitem{zhou2017scene}
B.~Zhou, H.~Zhao, X.~Puig, S.~Fidler, A.~Barriuso, and A.~Torralba.
\newblock Scene parsing through ade20k dataset.
\newblock In {\em CVPR}, 2017.

\bibitem{zhou2017voxelnet}
Y.~Zhou and O.~Tuzel.
\newblock Voxelnet: End-to-end learning for point cloud based 3d object
  detection.
\newblock {\em arXiv preprint arXiv:1711.06396}, 2017.

\end{thebibliography}
}

\end{document}